\documentclass[letterpaper, 10 pt, conference]{ieeeconf}
\IEEEoverridecommandlockouts 
\usepackage{algorithm}
\usepackage[noend]{algpseudocode}
\usepackage{array}
\usepackage{float}
\usepackage{graphicx}
\usepackage{amsmath}
\usepackage{multirow}
\usepackage{amssymb}
\usepackage{booktabs}
\usepackage{tikz}
\usepackage{ifthen}
\usepackage{blindtext}
\usepackage{siunitx}
\usetikzlibrary{positioning}
\usepackage{graphicx}
\usepackage{tikz}
\usepackage{amsmath}
\usepackage{xcolor}
\usetikzlibrary{positioning}

\tikzstyle{conv} = [rectangle, draw, fill=white!20,
text width=7em, text centered, minimum height=2em]
\tikzstyle{bn} = [rectangle, draw, fill=white!20,
text width=7em, text centered, minimum height=2em]
\tikzstyle{act} = [rectangle, draw, fill=white!20,
text width=7em, text centered, minimum height=2em]
\tikzstyle{concat} = [rectangle, draw, fill=white!20,
text width=7em, text centered, minimum height=2em]
\DeclareMathAlphabet\mathbfcal{OMS}{cmsy}{b}{n}
\usepackage[pagebackref,breaklinks,colorlinks]{hyperref}

\usepackage[capitalize]{cleveref}
\crefname{section}{Sec.}{Secs.}
\Crefname{section}{Section}{Sections}
\Crefname{table}{Table}{Tables}
\crefname{table}{Tab.}{Tabs.}

\title{\LARGE \bf
Point2Point : A Framework for Efficient Deep Learning on Hilbert sorted Point Clouds with applications in Spatio-Temporal Occupancy Prediction}

\author{Athrva Atul Pandhare$^{1}$
\thanks{$^{1}$The author is with the General Robotics, Automation, Sensing, \& Perception Lab (GRASP), University of Pennsylvania, PA 19104, USA.
        {\tt\small athrva@seas.upenn.edu}}%
}
\begin{document}
\maketitle

\begin{abstract}
The irregularity and permutation invariance of point cloud data pose challenges for effective learning. Conventional methods for addressing this issue involve converting raw point clouds to intermediate representations such as 3D voxel grids or range images. While such intermediate representations solve the problem of permutation invariance, they can result in significant loss of information. Approaches that do learn on raw point clouds either have trouble in resolving neighborhood relationships between points or are too complicated in their formulation. In this paper, we propose a novel approach to representing point clouds as a locality preserving 1D ordering induced by the Hilbert space-filling curve. We also introduce Point2Point, a neural architecture that can effectively learn on Hilbert-sorted point clouds. We show that Point2Point shows competitive performance on point cloud segmentation and generation tasks. Finally, we show the performance of Point2Point on Spatio-temporal Occupancy prediction from Point clouds. 
\end{abstract}

\section{Introduction}

Deep learning on point clouds has been an active research area in recent years. Existing literature usually converts point clouds to an intermediate representation before learning. However, such intermediate representations have their own disadvantages. Voxel grid-based methods suffer from space inefficiency and information loss, while range image-based methods may not be able to capture the full 3D information of the point clouds. Direct learning on raw point clouds is a more space-efficient approach, but it requires the neural network to address the permutation invariance and irregularity of point clouds.

In this paper, we propose a novel representation of point clouds in the form of a sorting order induced by the Hilbert space-filling curve. This representation preserves locality and allows for the use of operations like traditional convolutions directly on raw point clouds. We also introduce the use of the Sinkhorn distance as a loss function for generative deep learning on point clouds.

We present Point2Point, a 1D convolutional neural network designed for direct deep learning on raw point clouds. In this paper, we particularly emphasize the task of single-step occupancy prediction from raw point clouds. To that end, our proposed architecture is parameter-efficient and a step towards achieving real-time performance on point cloud-based map prediction tasks on compute-constrained hardware. We evaluate Point2Point variants, created based on the number of parameters, on single-step occupancy prediction tasks. We formulate the occupancy prediction task in three methodologies and compare the performance among different methodologies.

\section{Related Work}

Prior work on deep learning on point clouds has primarily focused on three approaches: voxel grid-based methods, range image-based methods, and direct learning on raw point clouds.

Voxel grid-based methods, such as those proposed in \cite{landing_zone, multiview_voxel, 3d_cvnet_obj_detection}, convert point clouds to a 3D voxel grid and use 3D convolutional neural networks for learning. Although effective, these methods suffer from space inefficiency and information loss due to the inherent sparsity of voxel grids.

Range image-based methods, such as those proposed in \cite{mersch, range_image2, range_image3}, convert point clouds to 2D range images before processing by standard 2D/3D CNNs. These methods may not be able to capture the full 3D information of point clouds.

Direct learning on raw point clouds is a more space-efficient approach that can handle higher resolution point clouds. PointNet \cite{pointnet}, one of the first neural networks capable of learning directly on point clouds, addresses the permutation invariance problem implicitly in network design. However, it does not explicitly model the local structure and interactions between points. Later work, such as PointNet++ \cite{pointnetpp} and PointCNN \cite{pointCNN}, propose hierarchical and convolutional approaches to address these issues.

Techniques that explicitly consider topology, such as FoldingNet \cite{foldingnet}, AtlasNet \cite{atlasnet}, and TearingNet \cite{tearingnet}, use topology for generative learning on point clouds. However, they may not be suitable for point clouds consisting of multiple disconnected objects. TearingNet proposes tearing-based deformations of underlying grids to model complex topologies containing multiple detached objects.

\subsection{Contributions}

The main contributions of our work are:

\begin{itemize}
\item We propose a novel representation of point clouds in the form of a sorting order induced by the Hilbert space-filling curve. This representation preserves the locality of the point cloud and allows for the use of traditional convolution operations directly on raw point clouds.
\item We introduce Point2Point, a 1D convolutional neural network architecture for point clouds. Our architecture is designed to be parameter efficient, which is particularly important for achieving real-time performance on compute-constrained hardware. We evaluate and compare our model with state of the art models in point cloud segmentation and reconstruction tasks.
\item We define and formulate the single-step occupancy prediction task in the form of three methodologies and perform a qualitative and quantitative comparison among different methodologies. Finally, we show that Point2Point produces promising results on single-step occupancy prediction.
\end{itemize}

\section{Sorting Induced by the Hilbert Space-Filling Curve}

Following \cite{hilbert_curve_locality}, we formally define a space filling curve as follows - We consider a mapping $f$ defined as,
\begin{equation}
    f : \mathcal{I} \to \mathbb{E}^n \;\;\;\;\; \forall n \geq 2
\end{equation}
where $\mathcal{I} = [0, 1]$ is a unit interval, and $\mathbb{E}^n$ is $n$ dimensional Euclidean space. If the image of the interval $\mathcal{I}$ under the mapping $f$, has positive Jordan content, then $f(\mathcal{I})$ is said to be a space-filling curve. For a Hilbert space filling curve, the mapping between the substructures of the interval $\mathcal{I}$ and the square $S$ is such that it preserves inclusion relationships \cite{hilbert_curve_locality}. \cite{3d_hilbert_curve} shows that the Hilbert curve is a continuous and surjective mapping. \cite{hcl_1,hcl_14} show that Hilbert curves have the best locality preserving properties among different space filling curves\cite{sagan_1994_spacefilling,hilbert_curve_locality,hcl_5,hcl_14,3d_hilbert_curve}. We use the Hilbert sorting algorithm to induce a locality preserving ordering over point clouds.
\subsection{The Hilbert Sorting Algorithm}
The Hilbert curve is a space-filling curve that maps multi-dimensional data to one dimension while preserving the locality of the data\cite{hilbert_curve_locality}\cite{3d_hilbert_curve}. In this section, we describe the algorithm for sorting a point cloud using the Hilbert curve induced ordering\cite{hilbert_sort_algo}\cite{hilbertalgo}.
Given a point cloud $P$ of size $n$, the Hilbert sorting algorithm is shown in \cref{hilbert_algorithm}:
\begin{algorithm}
\caption{Hilbert Curve Sorting Algorithm for $n$-Dimensional Point Set $P$}\label{hilbert}
\begin{algorithmic}[1]
\Procedure{HilbertSort}{$P$}
\State Compute the bounding box of $P$.
\State Choose a level $n$ such that the bounding box fits within a unit $n$-hypercube of the Hilbert curve of level $n$.
\State Map each $n$-dimensional point in $P$ to a one-dimensional index using the Hilbert curve mapping $H_n$.
\State Sort the one-dimensional indices using radix sort.
\State Traverse the sorted sequence of one-dimensional indices in order and map each index back to its original $n$-dimensional spatial coordinate using the inverse of the Hilbert curve mapping $H_n^{-1}$.
\EndProcedure
\end{algorithmic}
\label{hilbert_algorithm}
\end{algorithm}
The Hilbert curve mapping $H_n$ maps a spatial coordinate $(x_1,x_2, \dots, x_n)$ to a one-dimensional index $i$ by interleaving the bits of the binary representations of $x_i$. The inverse Hilbert curve mapping $H_n^{-1}$ computes the spatial coordinate from the one-dimensional index $i$. Both mappings are defined recursively based on the orientation of the subcurve within the unit square.

\section{Loss function : The Sinkhorn Distance} \label{sinkhorn_loss_algorithm}
The Sinkhorn Algorithm is an iterative algorithm used for approximating the Wasserstein distance between two probability distributions. It can be used as a loss function in generative tasks and has been shown to be more effective than the commonly used Chamfer distance\cite{sinkhorn_algorithm_supp}. The algorithm is based on an entropic regularization of the Wasserstein distance and is fully differentiable, making it well-suited for use in deep learning applications\cite{sinkhorn_algorithm_supp, sinkhorn_reference}.

Given two point clouds $K$ and $G$, represented as discrete measures, the Sinkhorn Algorithm solves the objective in Equation \ref{sinkhorn_objective}, which is $\epsilon$-strongly convex, and therefore has a unique optimal solution $P^*$.

\begin{equation}
L(K, G) = \min_{P} \langle P_{K,G}, C_{K,G} \rangle - \epsilon H(P_{K,G})
\label{sinkhorn_objective}
\end{equation}

Here, $C_{K,G}$ is the cost matrix associated with the two point clouds under an $L_n$ norm, and $H(P_{K,G})$ is the discrete entropy of the coupling matrix $P_{K,G}$, given by Equation \ref{discrete_entropy}.

\begin{equation}
H(P) = - \sum_{i,j} P_{i,j}(\log(P_{i,j}) - 1)
\label{discrete_entropy}
\end{equation}

The Sinkhorn Algorithm computes the optimal coupling matrix $P^*$ by iteratively updating the vectors $u$ and $v$ and the matrix $P$, as shown in Algorithm \ref{alg:sinkhorn_P}.
\begin{figure*}
  \centering \includegraphics[width=\textwidth]{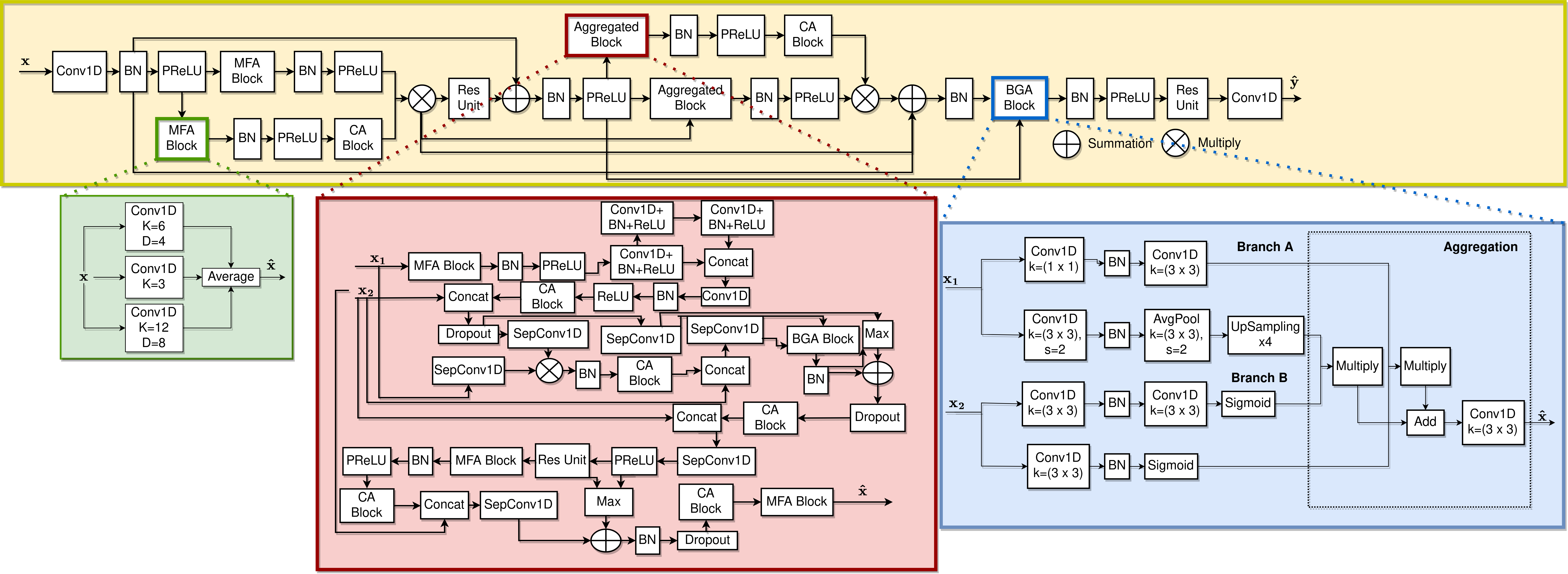}
  \caption{The complete architecture of Point2Point, including the Multiscale Feature Aggregation Block (MFA), the Bilateral Feature Aggregation Block (BFA), Channel Attention (CA) block and the Aggregated Block. The structure of the MFA, BFA and Aggregated blocks is shown in further detail. (All the Convolutional layers are \texttt{conv1D}, BN$\to$\texttt{BatchNorm}, ResUnit$\to$\texttt{Single Residual Convolution block}, SepConv$\to$\texttt{SeparableConvolution})}
  \label{complete_architecture}
\end{figure*}
\begin{algorithm}[h]
\caption{Sinkhorn Algorithm for Wasserstein Distance}
\label{alg:sinkhorn_P}
\begin{algorithmic}[1]
\State \textbf{Input:} Point clouds $X = {x_1, \dots, x_n}$ and $Y = {y_1, \dots, y_n}$, cost function $C$
\State \textbf{Output:} Wasserstein distance $W_p(X,Y)$
\State \textbf{Initialize:} $\lambda = \epsilon = 1$, $K = \exp(-C/\epsilon)$, $u = \frac{1}{n}\mathbf{1}$, $v = \frac{1}{n}\mathbf{1}$
\While{not converged}{
\State $u \leftarrow \frac{1}{Kn}$ \hspace{0.25cm} \textit{(Dimension: $n \times 1$)}
\For{$i = 1$ to $n$}
\State $u_i \leftarrow \frac{u_i}{\sum_{j=1}^n K_{ij}v_j}$ \hspace{0.25cm} \textit{(Dimension: $n \times 1$)}
\EndFor
\State $v \leftarrow \frac{1}{K^Tu}$ \hspace{0.25cm} \textit{(Dimension: $n \times 1$)}
\For{$j = 1$ to $n$}
\State $v_j \leftarrow \frac{v_j}{\sum_{i=1}^n K_{ij}u_i}$ \hspace{0.25cm} \textit{(Dimension: $n \times 1$)}
\EndFor
}
\EndWhile
\State $P^*$ = \texttt{diag}(u)K\texttt{diag}(v)
\State $W_p(X,Y) = \langle P^*, C \rangle - \epsilon H(P^*)$
\end{algorithmic}
\end{algorithm}

The Sinkhorn distance between the point clouds $K$ and $G$ is then calculated as shown in Equation \ref{sinkhorn_loss}, where $\langle .,. \rangle$ is the inner product.
\begin{equation}
d_{K,G} = \langle P_{K,G}, C_{K,G} \rangle - \epsilon H(P_{K,G})
\label{sinkhorn_loss}
\end{equation}
Overall, the Sinkhorn Algorithm is an attractive loss function for deep learning tasks due to its differentiability.

\section{Proposed Neural Architecture for Point Cloud Data}
We propose a neural network architecture designed specifically to learn from sorted point cloud data. The neural network consists of multiple building blocks that we explain in the following sections. The proposed architecture is a fully-1D convolutional network that utilizes 1D convolutional layers operating along the first dimension of a $\mathbb{R}^{n\times 3}$ dimensional point cloud. This enables the network to capture the dependencies among neighboring points. Since the point clouds are sorted using Hilbert curves, the convolution operation applies only to points that are spatially close, eliminating the need for nearest-neighbor search based ball querying to find neighborhoods. 

\subsection{Multi-Scale Feature Aggregation}
We propose the Multi-Scale Feature Aggregation (MFA) block, which estimates point relationships at varying scales and aggregates them to form a coherent representation. The MFA block consists of 1D convolutional layers with increasing kernel sizes and dilation rates, expanding the receptive field of the neural network. The information accumulated from these layers produces a multi-scale feature representation of the input feature map.

\subsection{Bilateral Feature Aggregation}
We adapted the Bilateral Feature Aggregation (BFA) block from the design of the Bilateral Segmentation Network in \cite{bisenet,bisenet2}. The BFA block aggregates information from two upstream feature maps such that structural and discriminating lower-level feature information is preserved.

\subsection{Attentive Rechecking}
We use the Channel Attention (CA) block proposed in \cite{squeeze_and_excite} extensively in our neural network. We introduce the Attentive Rechecking (AR) methodology, which uses CA for importance-based recombination. AR passes the input through two learning-based blocks in a branched manner, after which the output of one branch is activated by the CA block and recombined with the output of the other branch via an aggregation layer. We use AR at several scales in our neural network.

\subsection{Aggregated Block}
Finally, we define the Aggregated block, which is a composition of the aforementioned blocks using Attentive Rechecking. The Aggregated block receives two input feature-maps, where the features in one input are heavily processed by modules like the MFA block, and the features in the second input are lightly processed and combined into deeper representations of the first input.

\Cref{complete_architecture} shows the proposed architecture containing the aforementioned learning blocks. 
\section{Experimentation}

\subsection{Datasets}

In our experiments, we utilize five distinct datasets namely ShapeNet\cite{shapenet}, KIMO-5\cite{tearingnet}, KITTI\cite{kitti_raw}, S3DIS\cite{s3dis}, and Semantic3D\cite{semantic3d}. Our generative tasks employ the ShapeNet, KIMO-5, and KITTI-raw datasets while the S3DIS and Semantic3D datasets are used for point cloud segmentation tasks. The ShapeNet dataset focuses on single-object scenarios while the KIMO-5 dataset is a multi-object dataset with a KIMO-$n$ dataset containing $k \leq n^2$ objects. We selected these datasets to assess the model's performance on both single and multi-object generative tasks. For single-step occupancy prediction, we opted for the KITTI dataset, specifically the raw LiDAR sequences, as it is a real-world dataset well-suited for occupancy prediction tasks. Furthermore, we modify the S3DIS and Semantic3D datasets to make them more challenging. We randomly remove a certain proportion, $\beta$, of points from a subset of the original dataset. We repeat this process for different values of $\beta$, specifically $\beta={0.1,\dots, 0.6}$, and then combine the resulting datasets to create our final modified dataset. This allows us to gauge segmentation performance, both on dense and sparse point clouds. Our aim in utilizing these datasets is to demonstrate the model's effectiveness in both segmentation and generative tasks.
\begin{figure*}
  \centering \includegraphics[width=0.79\textwidth]{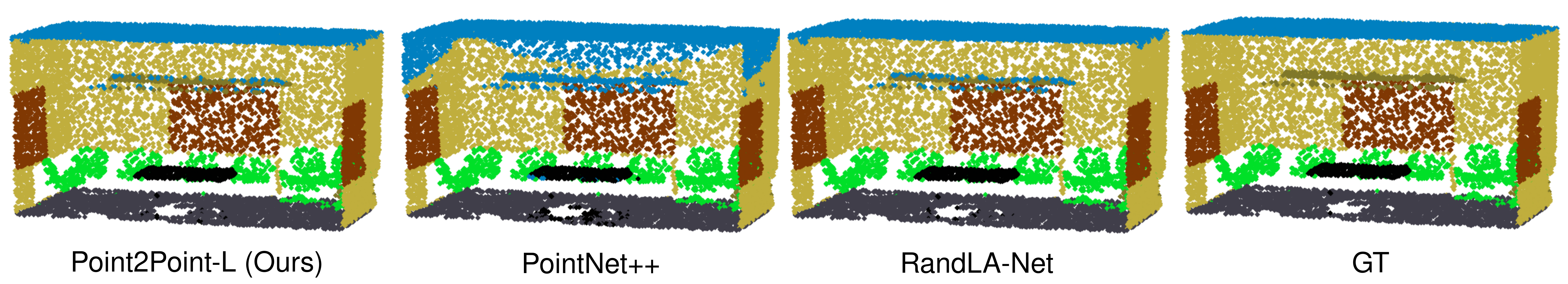}
    \caption{Results on Point cloud segmentation task. Point2Point-L variant was used to generate these results. It can be seen that Point2Point produces competitive results of high visual fidelity.}
  \label{segmentation_results}
\end{figure*}

\subsection{Implementation Details}
Our experiments are implemented using the TensorFlow framework \cite{tensorflow}. We use the Adam optimizer \cite{adam_optimizer} with a learning rate of \num{1e-3} and a batch size of 1. We use farthest point sampling for sub-sampling point clouds. For Occupancy prediction and Reconstruction, We train Point2Point for 450 epochs and run 175 Sinkhorn iterations with $\epsilon=0.001$ per sample. For segmentation, we train Point2Point for 150 epochs and use the cross-entropy loss function.

\subsection{Results on Point Cloud Segmentation}
We evaluate our proposed Point2Point model for point cloud segmentation tasks and compare its performance against several state-of-the-art models on two benchmark datasets: S3DIS\cite{s3dis} and Semantic3D\cite{semantic3d}. Table \ref{tab:segmentation_comparison_combined_9} presents the results of the comparison, where Point2Point achieves a competitive mIoU score of 76.1\% on modified S3DIS and 71.24\% on modified Semantic3D. \Cref{segmentation_results} shows a qualitative comparison between the performance of Point2Point compared with other architectures. 

\subsection{Results on Point Cloud Reconstruction}
We evaluate our proposed Point2Point model for point cloud reconstruction task and compare its performance against several state-of-the-art models on two benchmark datasets: ShapeNet and KIMO-5.
For a quantitative comparison, we track two metrics: Chamfer distance (CD) and Earth Mover's Distance (EMD). \Cref{tab1} shows the quantitative results of the comparison on the ShapeNet and KIMO-5 datasets. Point2Point outperforms all baseline models on both datasets, achieving the lowest CD and EMD scores. \Cref{reconstruction_exp} shows the qualitative results on the reconstruction task.

\begin{table}[h!]
\scriptsize
\centering
\makebox[\linewidth][c]{
\resizebox{0.7\columnwidth}{!}{
\begin{tabular}{*{1}{m{0.38\linewidth}}||*{1}{m{0.38\linewidth}}}

Input $K$& Reconstruction $\mathbf{\mathbf{h}}(K_i,\theta)$ \\
\hline
(a) \textbf{ShapeNet} & \\
    {\centering
  \includegraphics[width=0.9\linewidth]{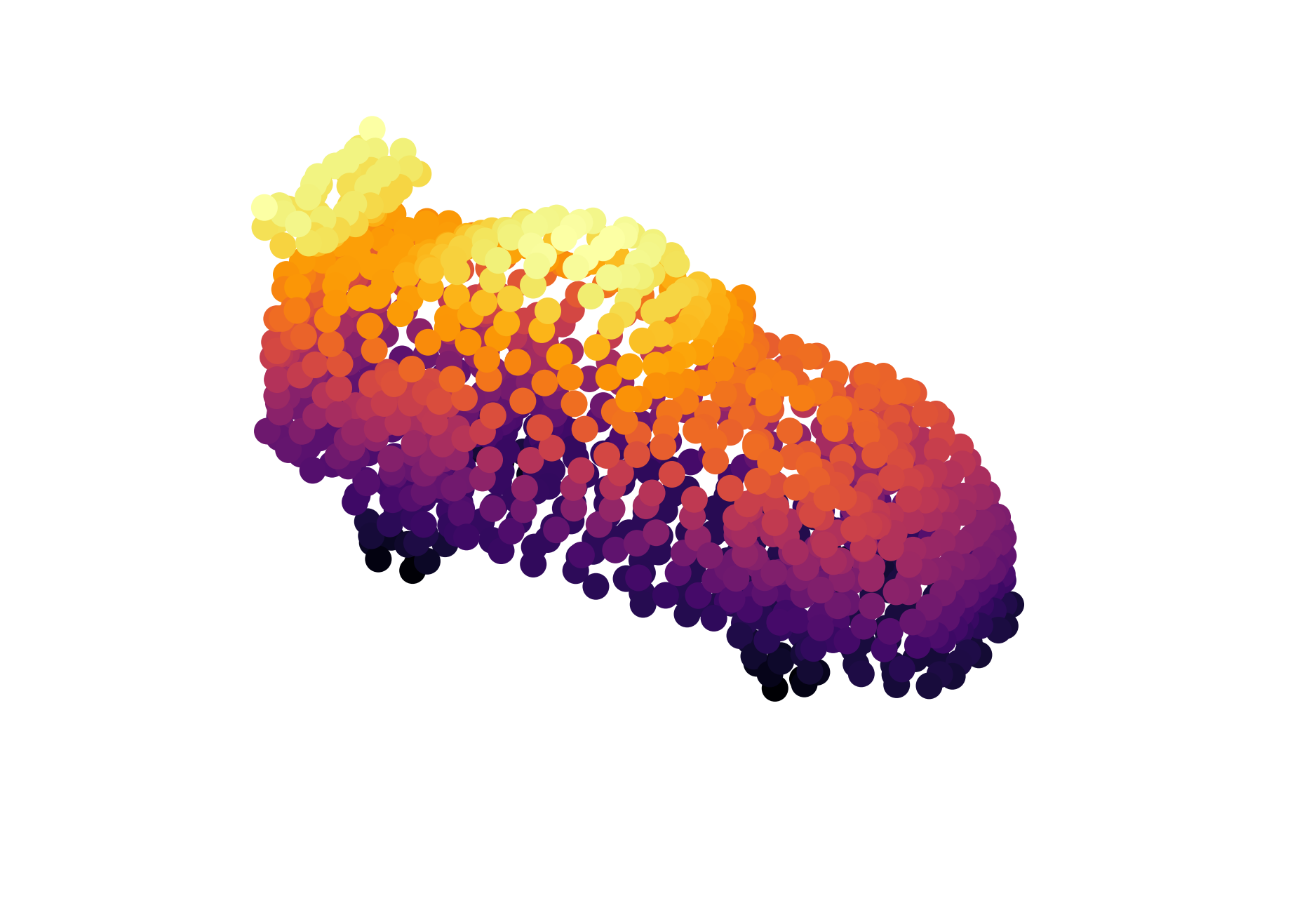} \par}
  &{\centering \includegraphics[width=0.9\linewidth]{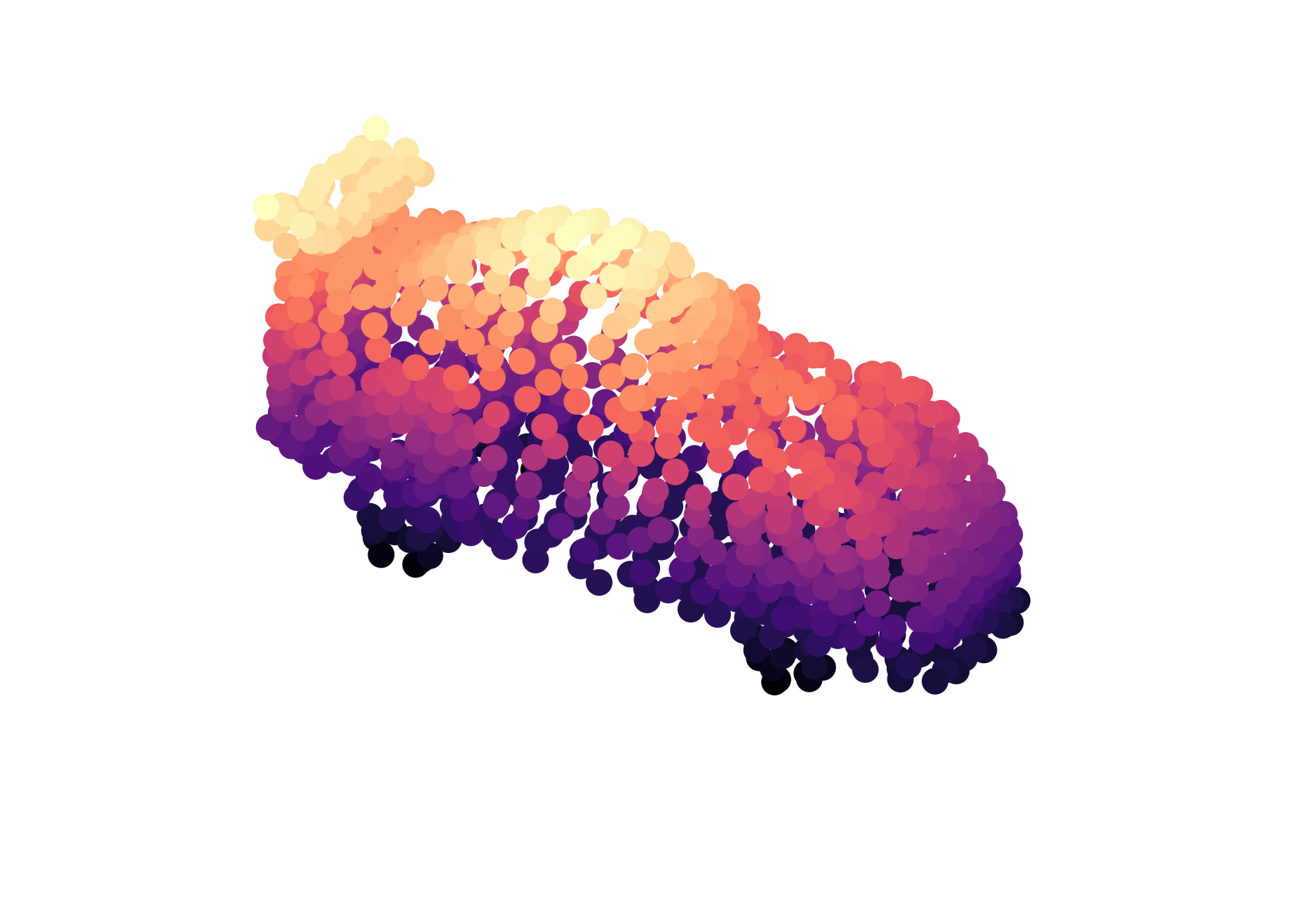} \par}\\

{\centering
  \includegraphics[width=0.9\linewidth]{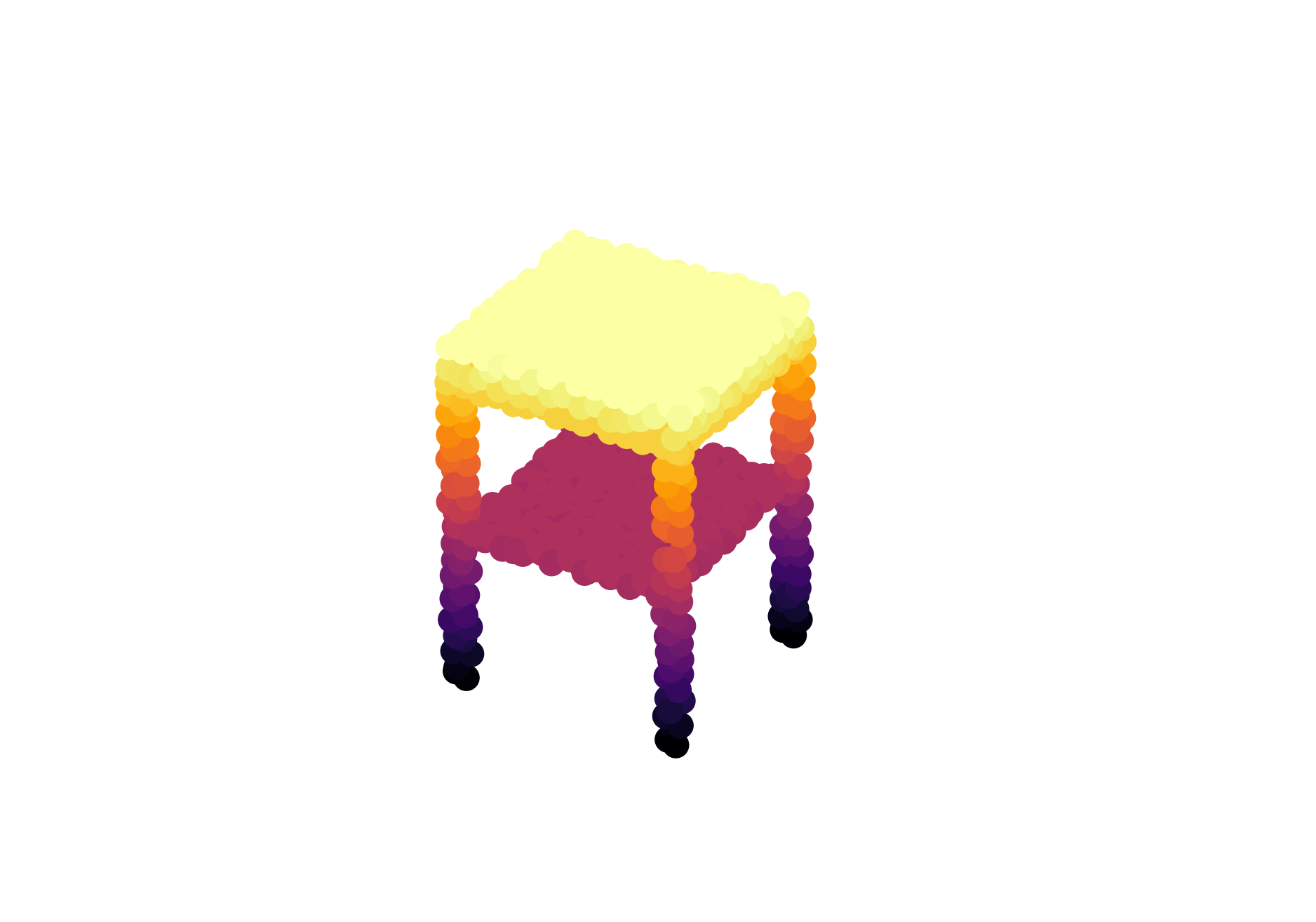} \par}
  &{\centering\includegraphics[width=0.9\linewidth]{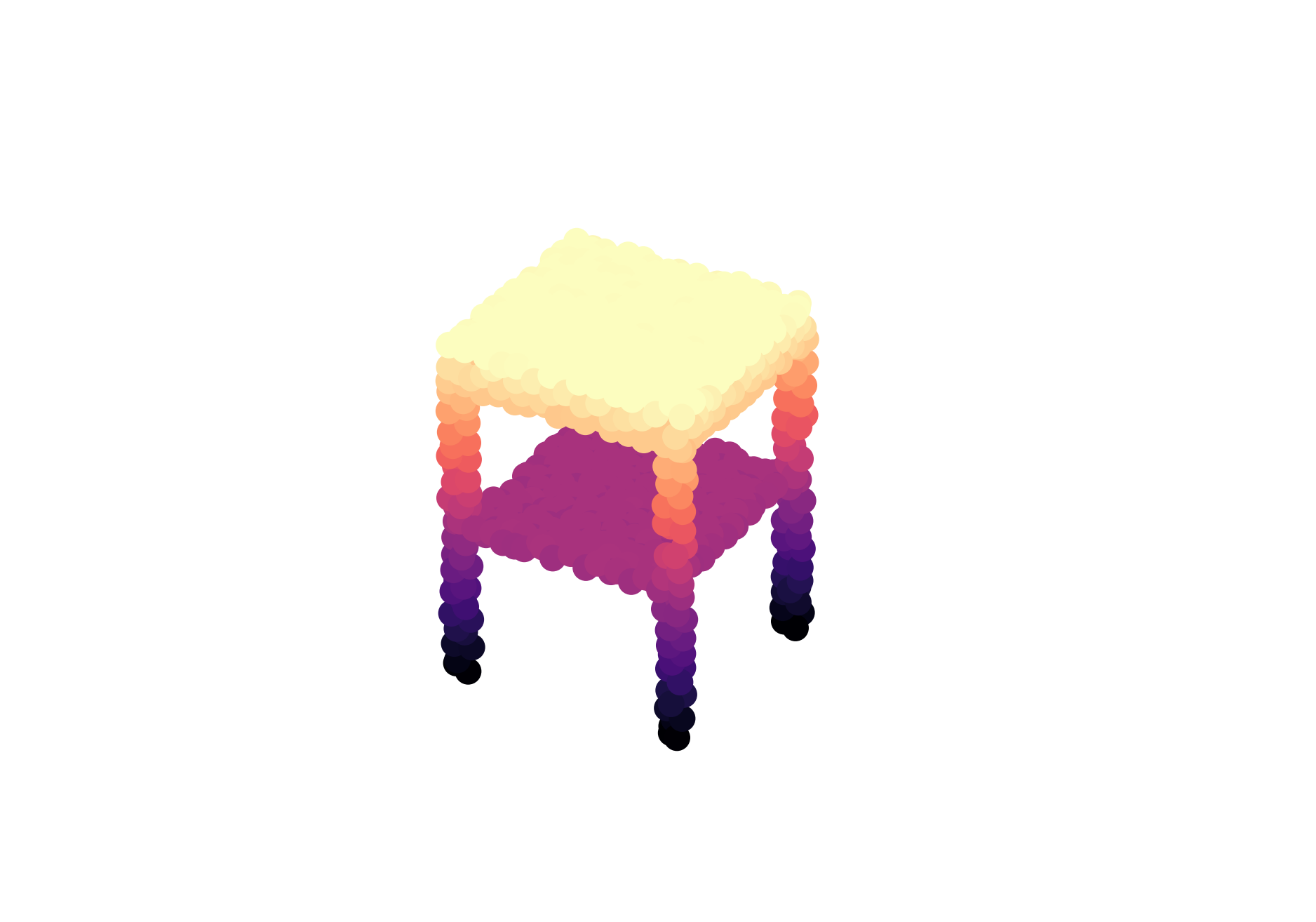} \par}\\
 \hline
 (b) \textbf{KIMO-5} with $k < 5^2$& \\
{\centering
  \includegraphics[width=0.9\linewidth]{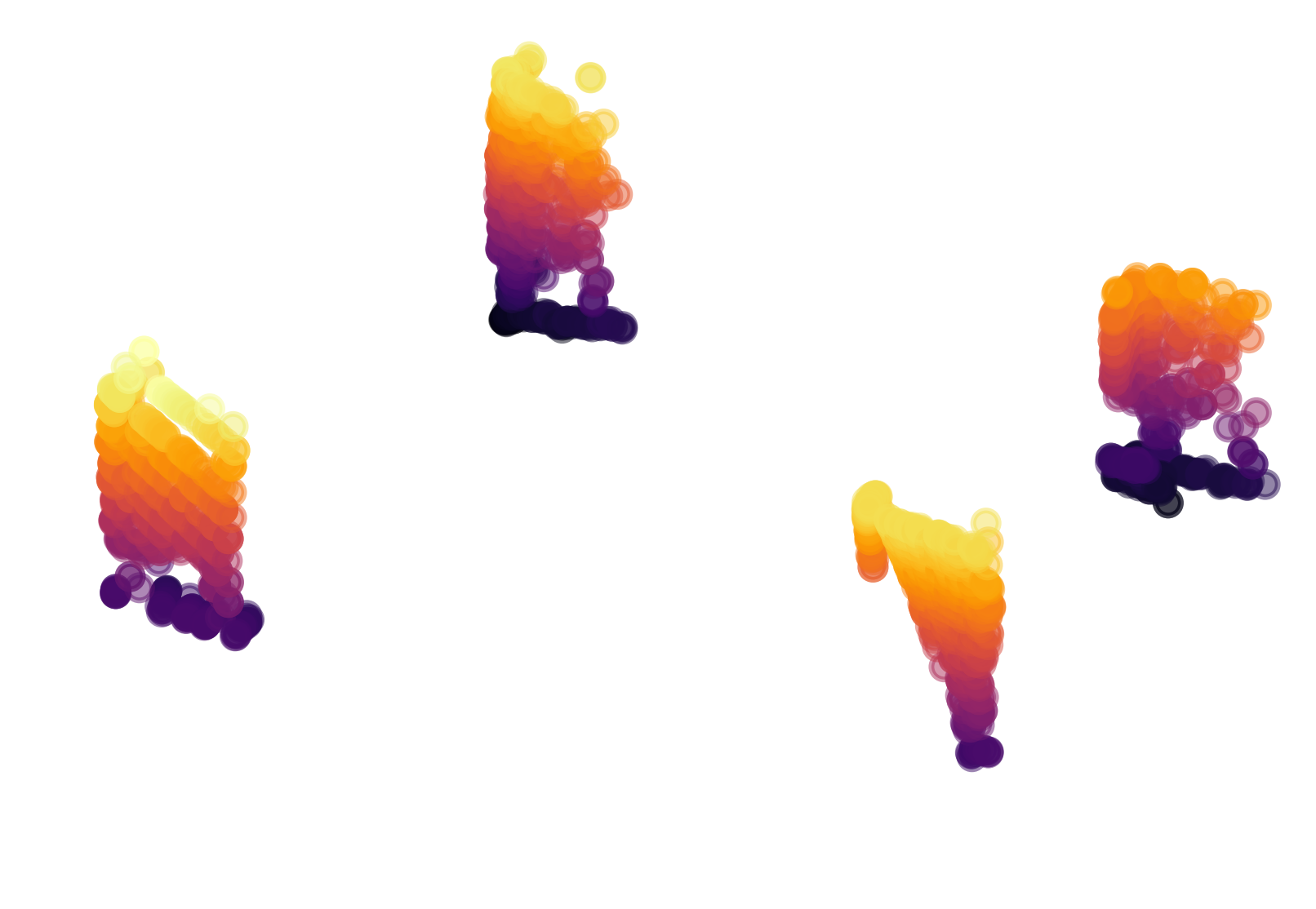} \par}
  &{\centering\includegraphics[width=0.9\linewidth]{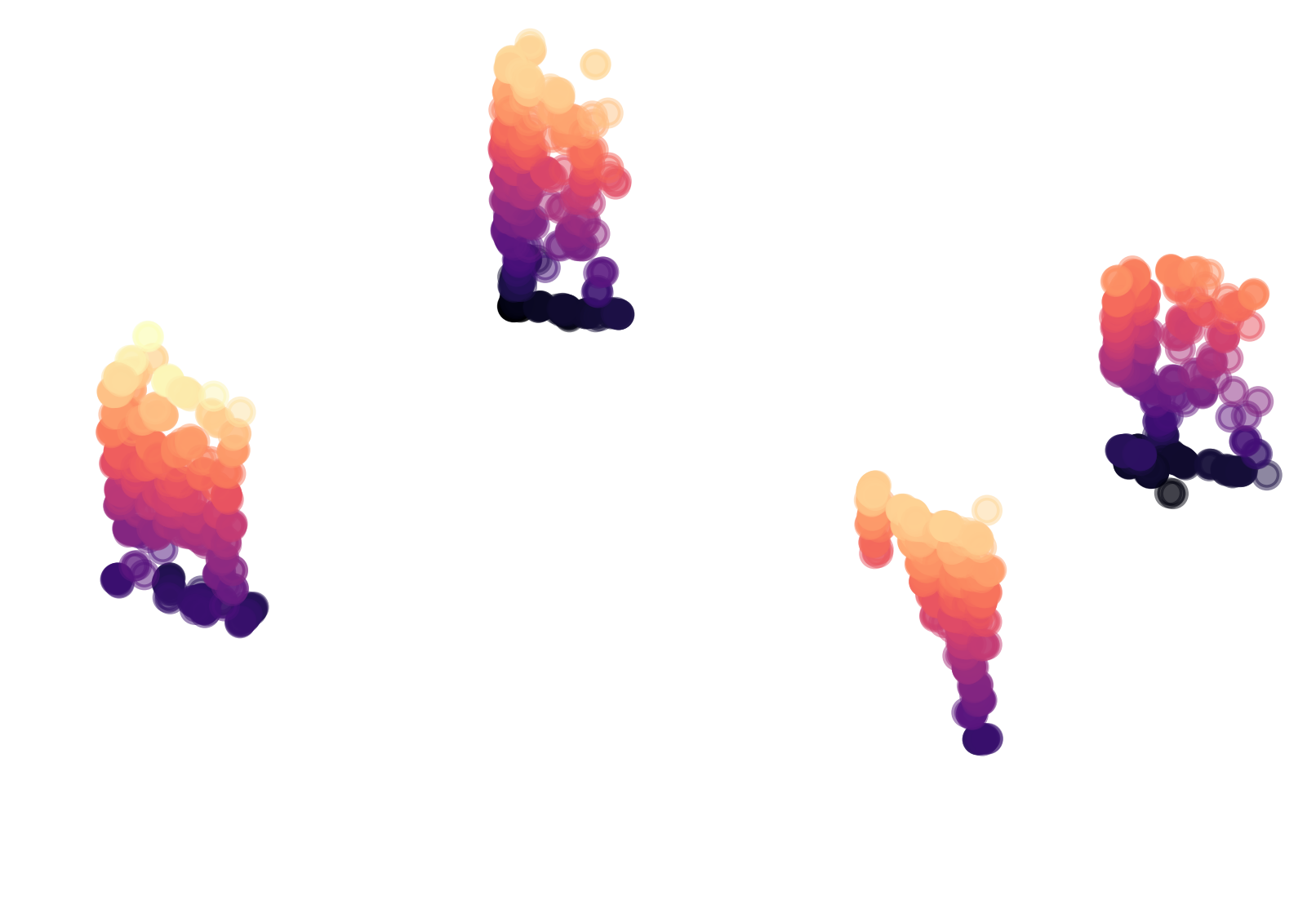} \par}\\
 \hline
 (c) \textbf{KIMO-5} with $k \approx 5^2$& \\
{\centering
  \includegraphics[width=0.9\linewidth]{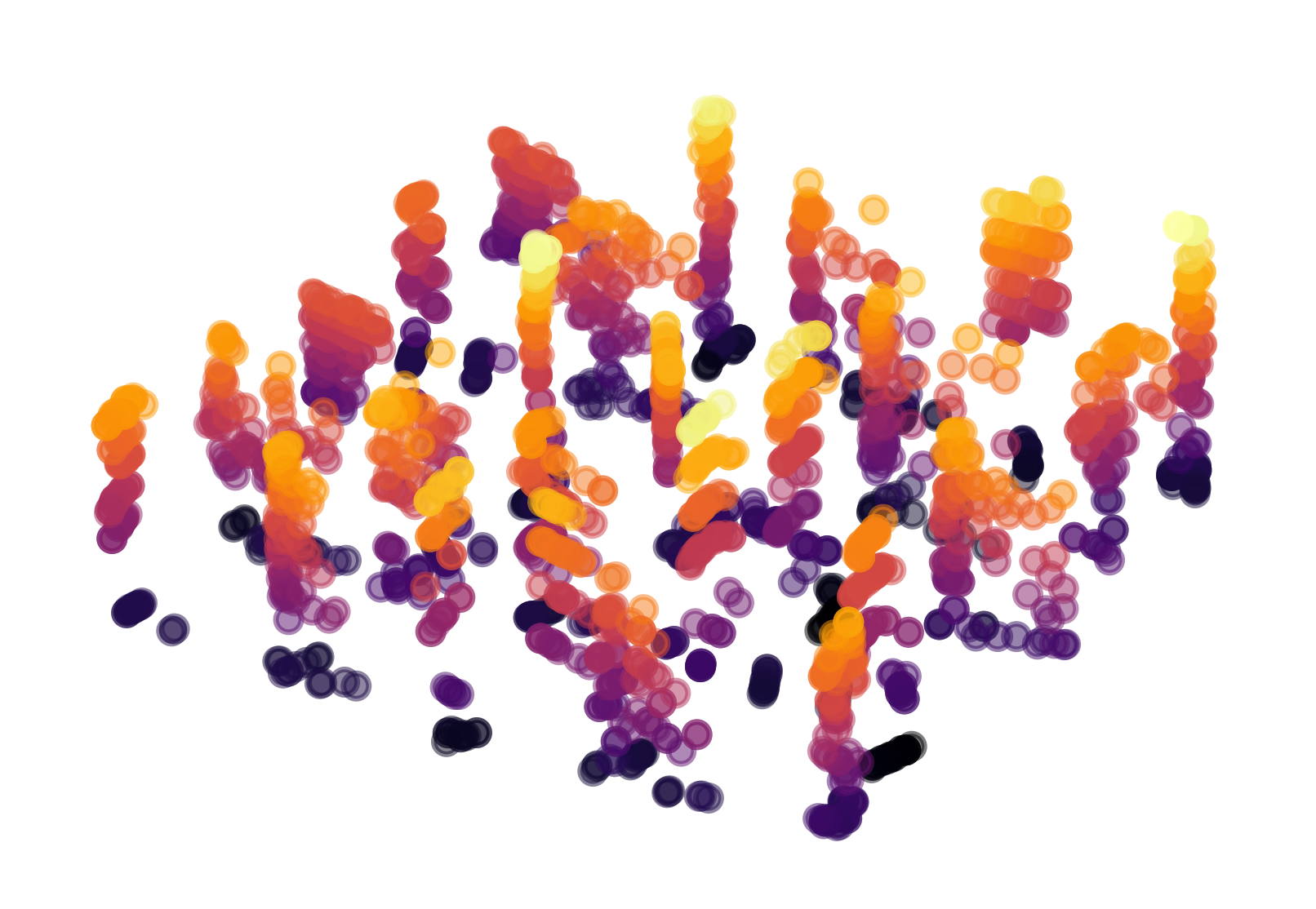} \par}
  &{\centering\includegraphics[width=0.9\linewidth]{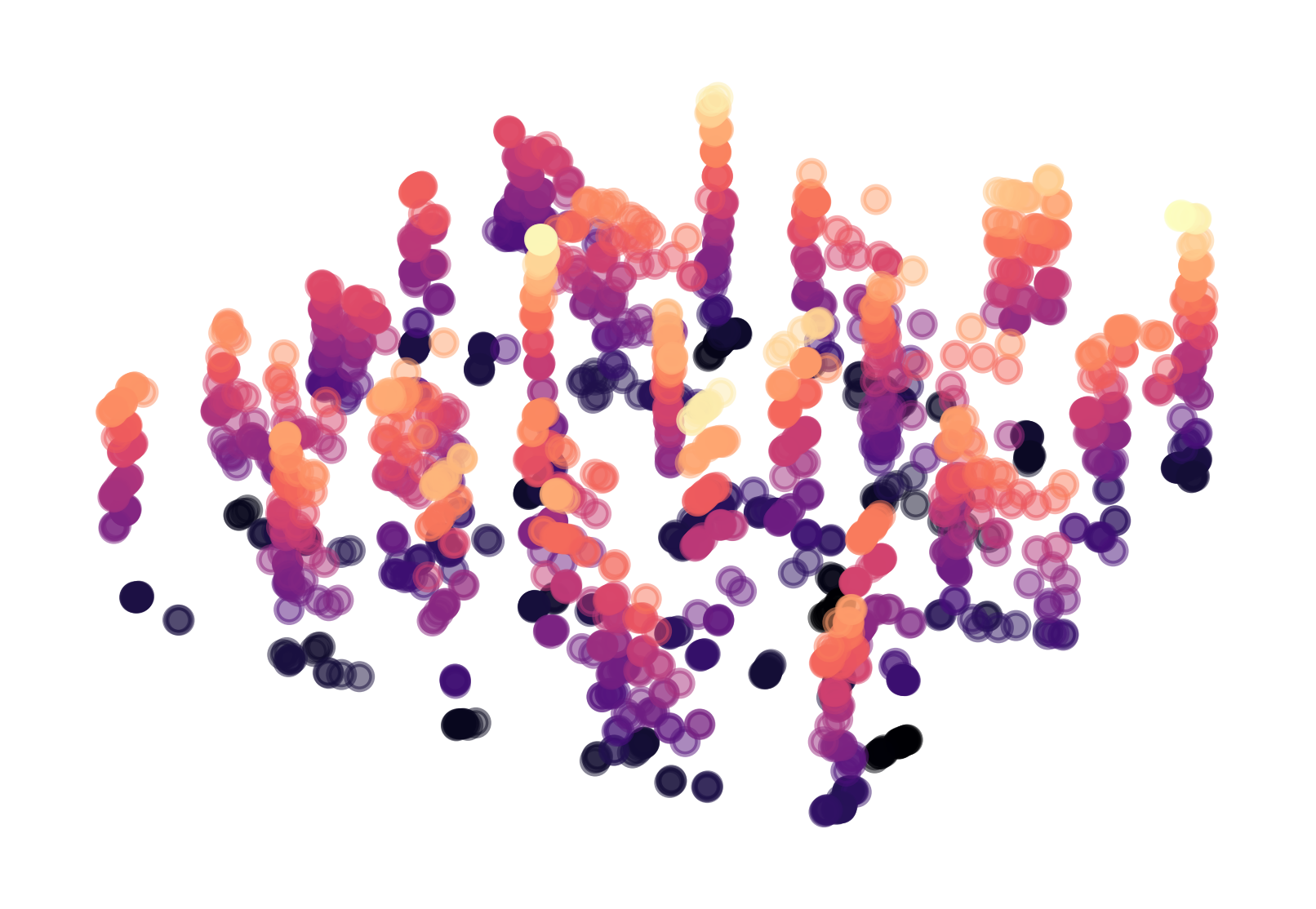} \par}\\
  \hline \\
\end{tabular}}}

\caption{Point cloud reconstruction results using Point2Point. (a) On Shapenet\cite{shapenet}, (b) On KIMO-5\cite{tearingnet} ($k$ $<$$5^2$), (c) On KIMO-5\cite{tearingnet} ($k$ $\approx$ $5^2$)}
\label{reconstruction_exp}
\end{table}

\begin{table}[htbp]
\caption{A Quantitative comparison of the Proposed Point2Point model to the Baseline architectures on Point cloud Reconstruction task.}
\centering
\resizebox{0.45\textwidth}{!}{
\begin{tabular}{ | c | c | c | c | c | }
\hline
\textbf{Metrics}&\multicolumn{2}{ | c | }{CD $\times 10^{-2}$} &\multicolumn{2}{ | c | }{EMD} \\
\cline{1-5} 
\textbf{Datasets} & \textbf{\textit{ShapeNet}}& \textbf{\textit{KIMO-5}}& \textbf{\textit{ShapeNet}}& \textbf{\textit{KIMO-5}} \\
\hline
LatentGAN\cite{latentgan} & 2.85& 17.18 & 0.218 & 3.773 \\
AtlasNet\cite{atlasnet}& 2.72&  9.59 & 0.163 & 3.173\\
FoldingNet\cite{foldingnet}& 2.75& 9.01 & 0.372 & 3.056 \\
Cascaded F-Net\cite{foldingnet}& 2.69& 9.13 & 0.207 & 2.944 \\
TearningNet\cite{tearingnet}& 2.54& 8.29 & 0.174 & 1.872 \\
TearingNet$_{3}$\cite{tearingnet}& 2.53&  8.24 & 0.169 & 1.867\\
\cline{1-5}
Point2Point-L(ours)& \textbf{2.37}& \textbf{8.11} & \textbf{0.143} & \textbf{1.572} \\
\hline
\end{tabular}}
\label{tab1}
\end{table}

\begin{table}[htbp]
\caption{A Quantitative comparison of the Proposed Point2Point model to the Baseline architectures on Point Cloud Segmentation task.}
\centering
\resizebox{0.35\textwidth}{!}{
\begin{tabular}{ | c | c | c |}
\hline
\textbf{Metrics}&\multicolumn{2}{ | c | }{mIoU (\%)} \\
\cline{1-3} 
\textbf{Models} & \textbf{\textit{S3DIS}}& \textbf{\textit{Semantic3D}}  \\
\cline{1-3} 
PointNet \cite{pointnet} & 57.9      & 53.8   \\
PointNet++ \cite{pointnetpp} & 69.9   & 62.2  \\
DGCNN \cite{DGCNN} & 73.5            & 65.1   \\
KPConv \cite{KPCONV} & 73.8          & 66.5  \\
RandLA-Net \cite{randla} & \textbf{76.6}       & 69.4   \\
PointCNN \cite{pointCNN} & 70.3       & 63.0  \\
ShellNet \cite{shellnet} & 74.5       & 66.4   \\
SalsaNet \cite{salsanet} & 69.8       & 63.3  \\
SpiderCNN \cite{spidercnn} & 72.1    & 64.7 \\
\cline{1-3}
Point2Point-L(ours)& 76.1& \textbf{71.2} \\
\hline
\end{tabular}}
\label{tab:segmentation_comparison_combined_9}
\end{table}

\section{Learning Single Step Spatio-temporal Occucpancy from 2D projected point clouds}

In this paper, we consider the task of occupancy prediction from point clouds, which is an important problem in autonomous driving. Specifically, we focus on predicting occupancy from 2D projected point clouds. We project the point clouds onto the $(x,y)$ plane and assume that the $z$-axis represents height. We use ground segmentation to remove redundant points and clip the point clouds so that only points within a distance of 30 meters from the ego vehicle in the $(x,y)$ directions are considered.

Our goal is to predict the expected occupancy of the environment in the next timestep, given the occupancy information in the current egocentric point cloud. This is known as the single-step occupancy prediction problem. We treat this as a supervised generative problem, where the input is the point cloud at time $t$ and the ground truth is the point cloud at time $t+1$. We seek to learn a model with weights denoted by $\theta$ that can estimate $\mathbf{x}_{t+1}$ given $\mathbf{x}_t$. 

We use three methodologies to approach the single-step occupancy prediction problem, which we present in the subsequent sections.

\subsection{Point cloud to Point cloud Learning}
\textit{Point cloud to Point cloud} learning concerns the estimation of the expected representation of a point cloud at time $t+1$, given a point cloud at time $t$. Specifically, let $\mathbf{x}_t$ denote a point cloud at time $t$, and let $\mathbf{x}_{t+1}$ denote the point cloud at time $t+1$ that we aim to estimate. Our objective is to approach this problem as a supervised generative task, where $\mathbf{x}_t$ serves as input and $\mathbf{x}_{t+1}$ is the ground truth we wish to predict. This task can be formulated as \cref{p2p2}, in which $\widehat{\mathbf{x}}_{t+1}$ is the predicted output and $\mathbf{h}(\mathbf{x}_{t},\theta)$ is a model parameterized by weights $\theta$.
\begin{equation}
     \widehat{\mathbf{x}}_{t+1} = \mathbf{h}(\mathbf{x}_t,\theta)
    \label{p2p2}
\end{equation}
We seek a model which minimizes the sinkhorn distance between all $ \widehat{\mathbf{x}}_{{t+1}_i}$ and $\mathbf{x}_{{t+1}_i}$ where $i=1,2,\dots, m$ is the number of sequence pairs. \Cref{p2p_problem}, shows such a model, where $P_{\mathbf{x}_{{t+1}_i}, \widehat{\mathbf{x}}_{{t+1}_i}}^*$ is the converged coupling matrix.
\begin{equation}
\begin{aligned}[b]
    \mathbf{h}^*(\mathbf{x}_{{t}_i},\theta) = \text{argmin}_{\mathbf{h}(\mathbf{x}_{{t}_i},\theta)} \langle P_{\mathbf{x}_{{t+1}_i}, \widehat{\mathbf{x}}_{{t+1}_i}}^*, C_{\mathbf{x}_{{t+1}_i}, \widehat{\mathbf{x}}_{{t+1}_i}} \rangle \\
    - \epsilon H(P_{\mathbf{x}_{{t+1}_i}, \widehat{\mathbf{x}}_{{t+1}_i}}^*)
\end{aligned}
    \label{p2p_problem}
\end{equation}

\subsection{Point cloud to Difference learning}
In this approach, the goal is to learn the difference between the point clouds at successive time-steps, given the current point cloud. This estimated difference is then incorporated into the current point cloud to construct the predicted point cloud. This technique bears similarity to the one proposed in \cite{Moharj} for occupancy grids. Let $\mathbf{x}_t$ denote a point cloud at time $t$, and $\mathbf{x}_{t+1}$ denote the point cloud at time $t+1$. Additionally, let $\mathbf{\Delta}_{t+1\vert t}$ be defined as the difference between the two point clouds $\mathbf{x}_{t+1}$ and $\mathbf{x}_{t}$, where $ \vert \mathbf{x}_{t+1} \vert = \vert \mathbf{x}_{t} \vert$ (as specified in \cref{diff_def_p2d}). Given $\mathbf{x}_{t}$, the objective is to predict $\mathbf{\Delta}_{t+1\vert t}$.

\begin{equation}
    \mathbf{\Delta}_{t+1\vert t} = \mathbf{x}_{t+1} - \mathbf{x}_{t}
    \label{diff_def_p2d}
\end{equation}
We now define $\mathbf{h}(\mathbf{x}_{t}, \theta)$ to be a model which learns the mapping in \cref{p2d_1}. 
\begin{equation}
    \widehat{\mathbf{\Delta}}_{t+1\vert t} = \mathbf{h}(\mathbf{x}_t, \theta)
    \label{p2d_1}
\end{equation}
The predicted point cloud $\widehat{\mathbf{x}}_{t+1}$ is obtained by \cref{p2d_2}.
\begin{equation}
     \widehat{\mathbf{x}}_{t+1} = \widehat{\mathbf{\Delta}}_{t+1 \vert t} + \mathbf{x}_t
     \label{p2d_2}
\end{equation}
This methodology provides a useful initial estimate ($\mathbf{x}_{t}$) for the prediction of ($\mathbf{x}_{t+1}$), i.e. it uses all the available information in $\mathbf{x}_{t}$ directly and requires the model $\mathbf{h}(\mathbf{x}_t, \theta)$ to only estimate the perturbations $\mathbf{\Delta}_{t+1\vert t}$. In other words, \textit{Point cloud to Point cloud} learning can be viewed as a two step process of first auto-encoding the input and then predicting the perturbations, whereas \textit{Point cloud to Difference} learning does not require auto-encoding and is only concerned with predicting the perturbations.
In a generalized sense, we seek a model for which satisfies \cref{p2p_problem}such that $\widehat{\mathbf{x}}_{{t+1}_i}$ is obtained from \cref{p2d_2}.
\begin{figure*}
  \centering \includegraphics[width=0.79\textwidth]{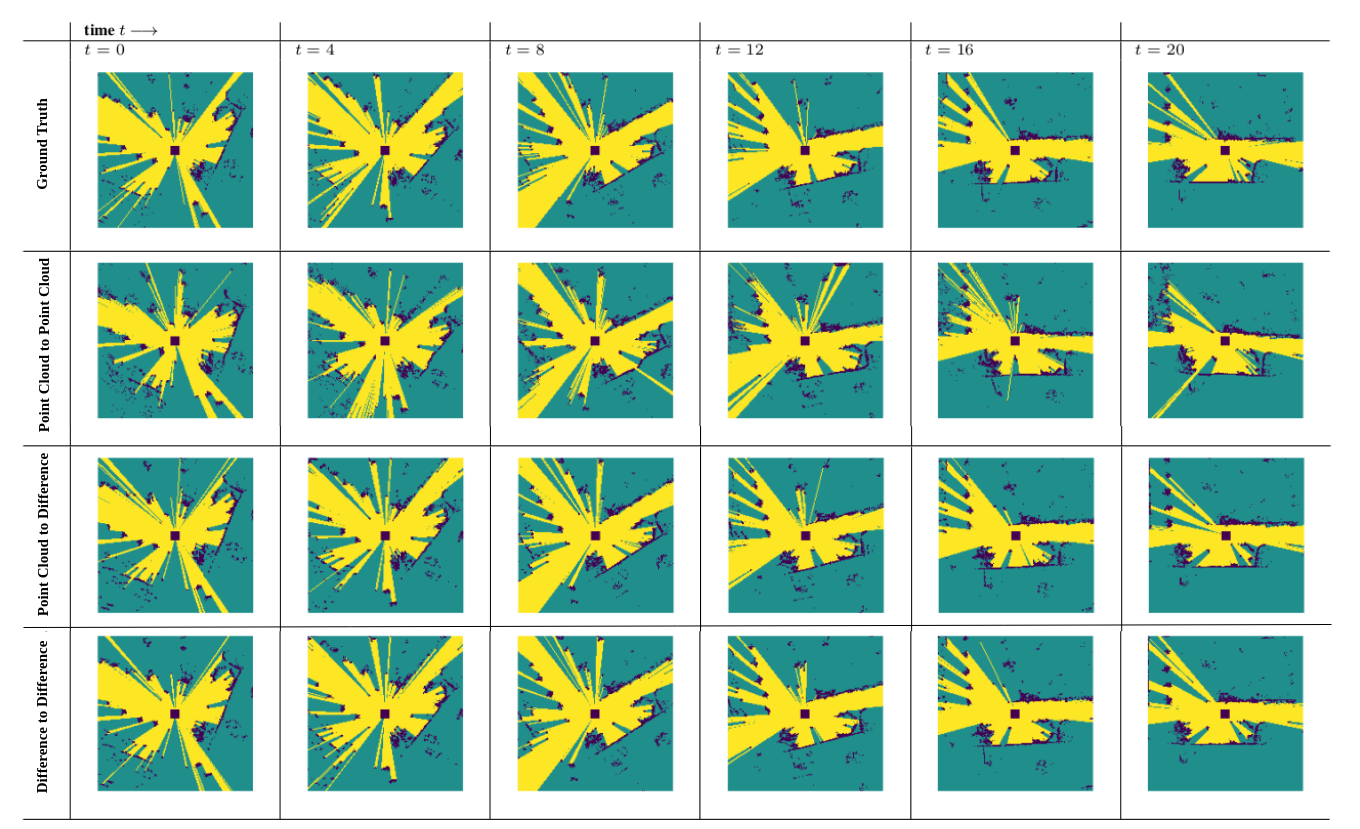}
    \caption{Results of Occupancy Prediction using (A) Point cloud to Point cloud, (B) Point cloud to Difference, (C) Difference to Difference Learning Methodologies. The raw point clouds are converted to occupancy grids for better visualization of differences between different methodologies. The visualizations are created using \textbf{Point2Point-L}.}
  \label{learning_results_occ_pred}
\end{figure*}
\begin{table*}[ht!]
\centering
\resizebox{0.7\textwidth}{!}{
\begin{tabular}{ | c | c | c | c | c | c | c | }
\hline
\textbf{Models}&\multicolumn{2}{|c|}{\textbf{A}}&\multicolumn{2}{|c|}{\textbf{B}}&\multicolumn{2}{|c|}{\textbf{C}}  \\
\cline{2-7}
 & CD $\times 10^{-2}$ & EMD$\times 10^{-2}$ & CD $\times 10^{-3}$ & EMD$\times 10^{-3}$ & CD $\times 10^{-3}$ & EMD$\times 10^{-2}$\\
\hline
\textbf{Point2Point-S} & 1.42 & 4.27 & 5.96 & 9.18 & 9.90 & 1.79\\
\textbf{Point2Point-M} & 1.28 & 2.92 & 4.03 & 7.92 & 8.17 & 1.51\\
\textbf{Point2Point-L} & \textbf{1.11} & \textbf{2.38} & \textbf{2.79} & \textbf{6.39} & \textbf{6.71} & \textbf{1.39}\\
\hline
\end{tabular}}
\caption{Performance of Point2Point variants on (\textbf{A}) Point cloud to Point cloud, (\textbf{B}) Point cloud to Difference, (\textbf{C}) Difference to Difference}
\label{table_occupancy_prediction}
\end{table*}
\subsection{Difference to Difference learning}

Difference to Difference (D2D) learning is a notion of learning difference between $\mathbf{x}_{t+1}$ and time $\mathbf{x}_{t}$, given the difference between $\mathbf{x}_{t}$ and time $\mathbf{x}_{t-1}$. It is an extension of the point cloud to difference methodology. It is identical to $\textit{Point cloud to Point cloud}$ with the exception that all the operations are on the differences $\mathbf{\Delta}_{t+1\vert t}$ and $\mathbf{\Delta}_{t\vert t-1}$. However, the implication of learning completely using $\mathbf{\Delta}_{t+1\vert t}$ and $\mathbf{\Delta}_{t\vert t-1}$ is worth considering. The difference $\mathbf{\Delta}_{t\vert t-1}$ encodes the change in the scene from $t-1$ to $t$. This information is essential in trying to estimate the motion of various objects. When this information is used to predict the change in scene from $t$ to $t+1$, some dynamic information is expected to be learned. 
\subsection{Results on Single Step Occupancy Prediction}
In this section, we evaluate the performance of three variants of Point2Point for occupancy prediction: Point2Point-S (0.253M parameters), Point2Point-M (0.894M parameters), and Point2Point-L (1.257M parameters). \textit{Note that for the segmentation and reconstruction tasks, the Point2Point-L variant was used}. It is worth mentioning that even the larger version, namely \textit{Point2Point-L}, comprises just 1.257M parameters.

Our evaluation results are presented in Table \ref{table_occupancy_prediction}. We find that the difference-based methodologies perform better than direct cloud to cloud translation, which is expected since they only require the model to predict the changes in the cloud from $t$ to $t+1$. Surprisingly, Point cloud to Difference learning outperforms Difference to Difference learning, and we attribute this to the approximation provided by the Sinkhorn algorithm for the true Wasserstein distance. Due to computational constraints, the algorithm may not iterate until convergence, resulting in a non-zero gradient and non-converging loss, especially for Difference to Difference learning, where the entries in $\mathbf{\Delta}_{t\vert t-1}$ and $\mathbf{\Delta}_{t+1\vert t}$ are small in magnitude and have a large overlap.
Qualitative comparisons of the proposed occupancy prediction methodologies are shown in \cref{learning_results_occ_pred}. We predict occupancy for five timesteps, where the cloud at $t-1$ is used to predict the cloud at $t$, and the true cloud at $t$ is used to predict $t+1$. Our results show that Point cloud to Difference learning produces the best results, followed by Difference to Difference learning, and Point cloud to Point cloud learning. We also observe that all learning methodologies produce better predictions closer to the ego vehicle than near the edges of the occupancy grids. The regions near the edges contain new information that cannot be accurately predicted, resulting in higher uncertainty. This suggests that additional context or temporal data aggregation may be required to improve performance in these regions.
\section{Ablation Study}
We incrementally remove the constituent blocks from \textbf{Point2Point-L} and train the resulting model on a subset of the ShapeNet\cite{shapenet} dataset on the reconstruction task. We then compare the resulting performance to the Performance of the original \textbf{Point2Point-L} model.
\begin{enumerate}
    \item \textbf{No} blocks removed $\longrightarrow$ $0.0$\%
    \item MFA Block (\textbf{1} instance removed)   $\longrightarrow$ $-3.62$\%
    \item MFA Block (\textbf{3} instances removed)   $\longrightarrow$ $-6.72$\%
    \item MFA Block (\textbf{all} instances removed) $\longrightarrow$ $-11.67$\%
    \item BFA Block (\textbf{all} instances removed) $\longrightarrow$ $-5.97$\%
    \item CA Block (\textbf{all} instancess removed) $\longrightarrow$ $-4.18$\% \\
\end{enumerate}
\section{conclusion}
This study introduces a new representation of point clouds using the Hilbert space-filling curve to create a locality preserving ordering. We propose Point2Point, a 1D convolutional neural network designed to operate on Hilbert-sorted point clouds. Our experiments on point cloud segmentation and reconstruction demonstrate that Point2Point performs competitively and outperforms baseline architectures, thus validating the effectiveness of our approach. We also propose three methodologies for addressing the single-step occupancy prediction task and show that Point2Point can be utilized for this task. Our experiments yield promising results, indicating the potential of Point2Point for real-time occupancy prediction on hardware with limited computational resources.
\newpage

{\small
\bibliographystyle{ieeefullname}
\bibliography{mainv2}
}

\end{document}